\pdfoutput=1

\documentclass[11pt]{article}

\usepackage{naacl2021}

\usepackage{times}
\usepackage{latexsym}

\usepackage[T1]{fontenc}

\usepackage[utf8]{inputenc}

\usepackage{subcaption}
\usepackage{xcolor}
\usepackage{amsmath}
\usepackage{algorithm2e}

\usepackage{multirow}
\usepackage{booktabs}
\usepackage[colorinlistoftodos,prependcaption,textsize=tiny]{todonotes}
\usepackage{xspace}
\usepackage{xargs} 
\newcommandx{\siva}[2][1=]{\todo[linecolor=red,backgroundcolor=red!10,bordercolor=red,#1]{SR: #2}\xspace}
\newcommandx{\yikang}[2][1=]{\todo[linecolor=black,backgroundcolor=gray!10,bordercolor=black,#1]{Yikang: #2}\xspace}
\newcommandx{\alex}[2][1=]{\todo[linecolor=green,backgroundcolor=green!10,bordercolor=green,#1]{ALEX: #2}\xspace}
\newcommandx{\aaron}[2][1=]{\todo[linecolor=green,backgroundcolor=blue!10,bordercolor=blue,#1]{AC: #2}\xspace}

\newcommand{\ignore}[1]{~}

\usetikzlibrary{arrows}
\usetikzlibrary{shapes}

\newcommand*\circled[1]{\tikz[baseline=(char.base)]{
            \node[shape=circle,draw,inner sep=1pt] (char) {#1};}}

\newcommand{\cell}[0]{\mathrm{cell}}
\newcommand{\cp}[0]{\overrightarrow{\pi}}

\newcommand{\q}[0]{{p'}}

\usepackage{microtype}

\title{Explicitly Modeling Syntax in Language Models \\
with Incremental Parsing and a Dynamic Oracle}

\author{Yikang Shen \\
  Mila/Universit\'e de Montr\'eal \\ 
  \texttt{yi-kang.shen@umontreal.ca}
  \And
  Shawn Tan \\
  Mila/Universit\'e de Montr\'eal \\
  \texttt{tanjings@mila.quebec} \\
  \AND
  Alessandro Sordoni \\
  Microsoft Research Montr\'eal \\
  \And
  Siva Reddy \\
  Mila/McGill University \\
  \And
  Aaron Courville \\
  Mila/Universit\'e de Montr\'eal \\
}

\date{}

\begin{document}

\maketitle
\begin{abstract}
Syntax is fundamental to our thinking about language.
Failing to capture the structure of input language could lead to generalization problems and over-parametrization.
In the present work, we propose a new syntax-aware language model: Syntactic Ordered Memory (SOM).
The model explicitly models the structure with an incremental parser and maintains the conditional probability setting of a standard language model  (left-to-right).
To train the incremental parser and avoid exposure bias, we also propose a novel dynamic oracle, so that SOM is more robust to wrong parsing decisions.
Experiments show that SOM can achieve strong results in language modeling, incremental parsing and syntactic generalization tests, while using fewer parameters than other models.
\end{abstract}

\section{Introduction}

Several recent works have systematically studied the linguistic abilities of modern language models, particularly syntax  \cite{linzen2016assessing,marvin2018targeted,gulordava2018colorless}.
They find that most language models are good at capturing frequent syntactic structures but do not generalize well to those in the long tail. Moreover, although some excel at having low perplexity scores, this is less due to their syntactic ability but more due to capturing collocations (frequently co-occurring words). Recently,~\citet{hu2020systematic} show that RNNs underperform on a syntactic generalization (SG) test set, whereas models that have an explicit notion of syntax, such as RNNG~\citep{dyer2016recurrent}, fare well on SG but at the cost of generally poorer language modeling (higher perplexity). Transformer-based models achieve strong performance when trained with large datasets, but are worse than random when trained on a small dataset.

\begin{figure}
    \centering
    \includegraphics[width=1\linewidth]{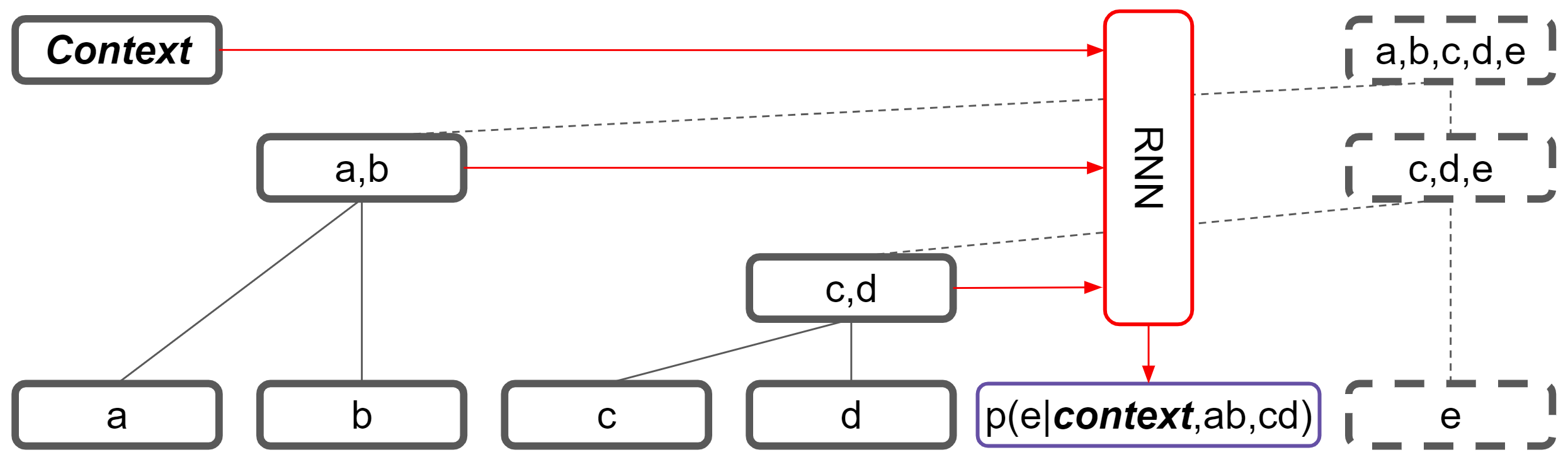}
    \caption{The mechanism of SOM.
    ``\textit{Context}'' is a distributed representation of previous sentences. 
    It could also represents the source sentence in a sequence to sequence task.
    It incrementally build subtrees given the input sentences. 
    A RNN will takes the context representation and the representations of subtrees in the current sentence to predict next token.
    }
    \vspace{-0.5cm}
    \label{fig:high_level}
\end{figure}

These works showed that building language models with an explicit internal model of syntax helps in achieving better performance in SG tasks and is also thought to help learn more efficiently in low data settings. However, building syntax-aware models that also obtain strong language modeling performance, when compared with recent transformer-based models, has until now seemed elusive. In this work, we propose a new syntax-aware language model dubbed Syntactic Ordered Memory (SOM; Fig.~\ref{fig:high_level}), which jointly acts as a language model and an \emph{incremental} parser. SOM inherits the syntax representation used in Ordered Memory (OM; ~\citealt{shen2019ordered}) in which syntax trees are embedded in a grid-like memory representation.
Whereas OM was trained as an unsupervised parser, SOM is explicitly trained to predict both ground-truth syntax trees incrementally and, using the predicted partial syntactic structure, to predict the next token. Fig.\ref{fig:high_level} shows the mechanism of SOM.

SOM factorizes the next-token prediction process into two steps: first, we predict the attachment position for the next token with a zero-step look-ahead parser, trained in a supervised fashion; then, we predict the next token distribution conditioned on the partially predicted structure. One way of training the incremental parser is to use teacher-forcing. However, this can lead to \emph{exposure bias}, due to the fact that the model was never exposed to its own predictions during training. To avoid this, we introduce a dynamic oracle~\citep{goldberg2012dynamic} for our model, so that our model can learn to recover from previous parsing mistakes during inference. We found this to be crucial to obtain good performance.

We compare SOM with existing methods that integrate syntax into language models.
RNNGs~\citep{dyer2016recurrent} and Ordered Neurons~\citep{shen2018ordered} are particularly related.
RNNGs are generative models of language which define a joint distribution on syntactic structures and sequence of words.
Ordered Neurons attempt to model the hierarchical structure of language by defining an ordering to the hidden states and the gates that impose that structure.
We show that our proposed SOM model can achieve strong language modeling, parsing and SG performance even when trained on small amounts of data.

In summary, our contributions are threefold:
\begin{itemize}
\item We introduce SOM, a new syntax-augmented language model that learns an incremental parser and use its predictions to improve language modeling.
\item We propose a novel dynamic oracle that allows to reduce the exposure bias and is instrumental to achieving good downstream performance.
\item We report high SG score, language modeling and incremental parsing performance for various dataset sizes. 
We also find that jointly learning both language modelling and parsing improves both these capabilities in the model.
\end{itemize}


\section{Related Work}\label{sec:related}

\paragraph{Syntax-aware models}
There has been work to integrate syntax into our current models of language.
\citet{socher2013recursive} used parse trees for composing sentences in order to predict sentiment over movie reviews.
However, having an external parser and restriction of batched computations in that early model made the method unwieldy.
\citet{bowman2016fast} introduced the SPINN model, which alleviated those issues, turning sentences into a sequence of actions to be executed by a shift-reduce parser.
Our SOM model is based on shift-reduce as well, because of the incremental nature of the parsing we want to achieve.
RNNG \cite{dyer2016recurrent,kuncoro2016recurrent} was an example of integrating syntax information for language modelling. 

There is also work that attempts to learn these syntactic structures without supervision.
\citet{kim2019unsupervised} later devised an unsupervised version of the RNNG, a method which produced good parsing performance.
DIORA \cite{drozdov2019unsupervised,drozdov2020unsupervised} was a method that leveraged the Inside-Outside algorithm to construct sentence embeddings for downstream tasks, with the benefit of being able to read off parse trees in the encoding process.

\citet{swayamdipta2019shallow} finds that there are no improvements over using ELMo \cite{peters2018deep} embeddings when shallow syntactic information is included, concluding that ELMo-style pretraining has learned the syntactic information.
However, \citet{kuncoro2019scalable} investigated the importance of the learnt syntactic knowledge RNNG in a large pre-trained model like BERT, they found that syntax information helps with downstream tasks.
In our experiments, we find that explicitly training OM with syntax (with our dynamic oracle scheme) improves performance on syntactic generalization tasks.

\paragraph{Incremental Parsing \& Language Modelling}
In SOM, we specifically focus on incremental parsing.
\citet{ghezzi1979incremental} discusses incremental parsing in the context of programming languages, with shift-reduce parsers being a specific type of incremental parsing.
OM, RNNG, and SPINN are models that were designed with shift-reduce in mind.

Incremental parsing lends itself well to the task of autoregressive language modelling.
Since the parser only sees the prefix of a sentence, the model can use the partial parse to make a prediction about upcoming words.
\citet{demberg2013incremental} summarises several empirical results that provide evidence for incremental and predictive parsing in humans, and makes several connections between \emph{incrementality} (that comprehenders do not wait to the end of the sentence before building a representation) and \emph{prediction} about future words coming in the sentence.

Given that an incremental parser processes a sentence from left to right, there are naturally some limitations.
\citet{hassan2009lexicalized} show why either a beam or delay is necessary if performing incremental parsing with monotonic extensions: They experiment with a parser based on Combinatory Categorial Grammar \cite{steedman2000syntactic}.
They find that without the look-ahead, there is a 30 \% point reduction in the parsing results.
One of our contributions in this paper is the one-step lookahead while performing parsing, but zero-step lookahead when performing next-word prediction, allowing the model to be trained jointly as a incremental parser and language model.

Despite the left-to-right nature of incremental parsing, this setting may aid language modelling too.
\citet{shieber1983sentence} suggests the biases may correspond to the way humans parse English, and use a modified shift-reduce parser to disambiguate between different parses of a sentence.
There have been work that show that incremental parsing can improve language modelling.
\citet{kohn2016predictive} demonstrate that combining an incremental dependency parser with a language model yields improvements in perplexity.
\citet{roark2001probabilistic} presents a top-down phrase structure parser that performs beam-search to generate connected intermediate structures for every sentence prefix.
This model can be used for language modeling and beats trigram models on the Penn Treebank \cite{marcus1994penn}

\paragraph{Dynamic Oracles}
Since incremental parsing requires that we break down the problem of structure prediction into sequential decisions, we are prone to \emph{exposure bias}.
There are techniques to address this by allowing the model to make mistakes and supervising future actions based on the state arrived at \cite{daume2009search}. 
\citet{goldberg2012dynamic} introduces the concept of dynamic oracles for dependency parsing.
\citet{coavoux2016neural} uses this technique for incremental constituency parsing, but uses morphological features, and does not perform language modelling.
\citet{fried2018policy} cover in further detail the related work relating to dynamic oracles and parsing.
We find that using dynamic oracles for training is crucial in seeing benefits in both language modelling and incremental parsing.
 
\paragraph{Evaluating Syntactic Generalization}

Recent tests have been developed that attempt to probe the linguistic abilities of language models.
\citet{gulordava2018colorless} explores the extent to which RNNs are able to model grammar, independent of the semantics of the sentence.
\citet{marvin2018targeted} evaluate language models on their ability to score sentences with and without the proper subject-verb agreements over a variety of different settings.

\citet{hu2020systematic} expands on these ideas, and propose a suite of syntactic generalization tests for language models over a series of different sized datasets.
They find that while GPT-2 performs well, their performance is highly dependent on the scale of the language modeling training dataset, while other models remain more robust. 
In this paper, we use this test suite for the evaluation.

\section{Ordered Memory}

\begin{figure}
    \centering
  \includegraphics[width=1\linewidth]{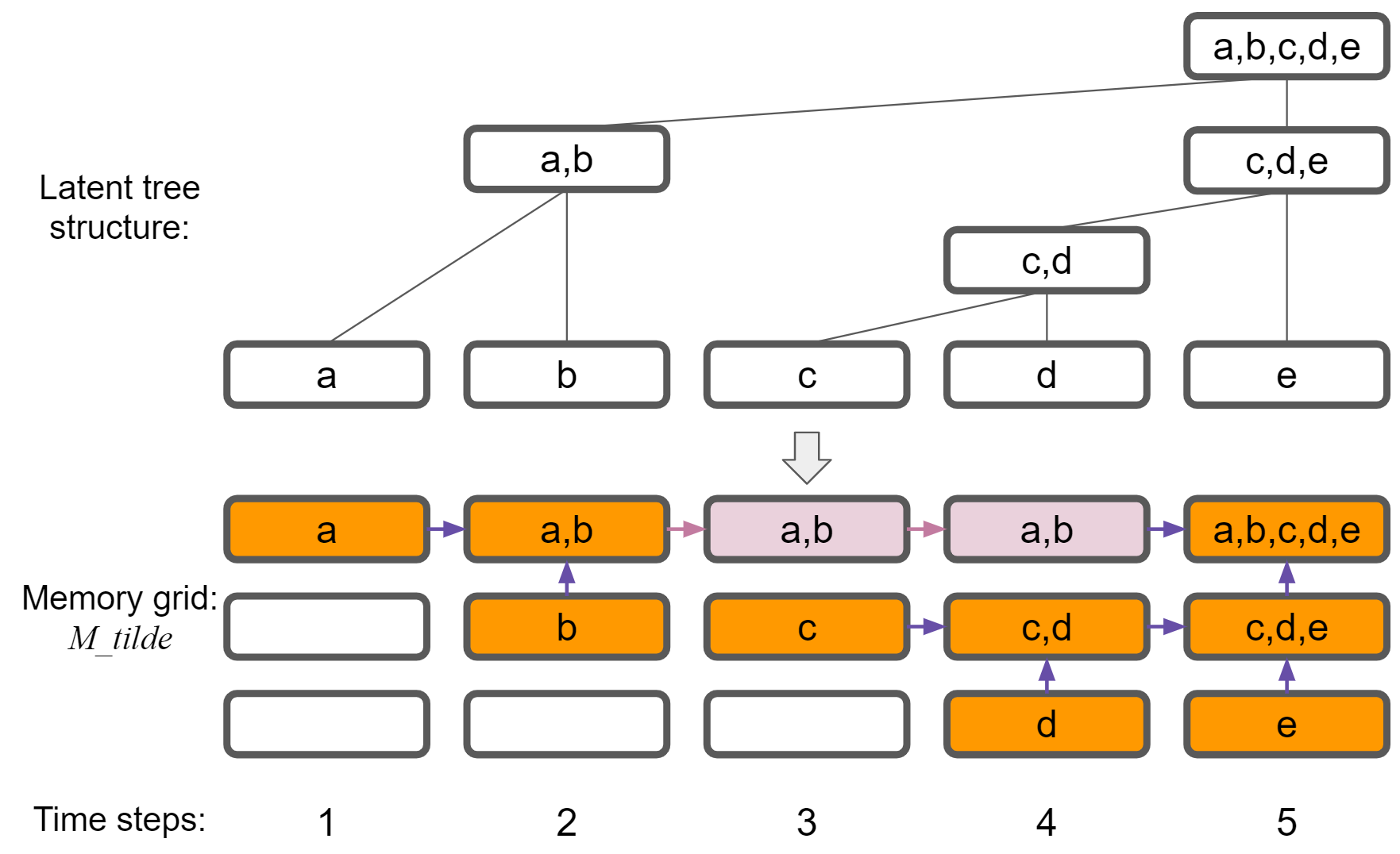}
  \caption{ The grid view of a tree structure. Blue arrows represent composing children into parent. Gray arrows represent copying from previous time step. Orange slots are memories generated at the current time step. Gray slots are memories copied from previous time step.}
  \vspace{-0.3cm}
  \label{fig:grid_view}
\end{figure}

\begin{figure*}
\centering
\begin{subfigure}{.5\textwidth}
  \centering
  \includegraphics[width=1\linewidth]{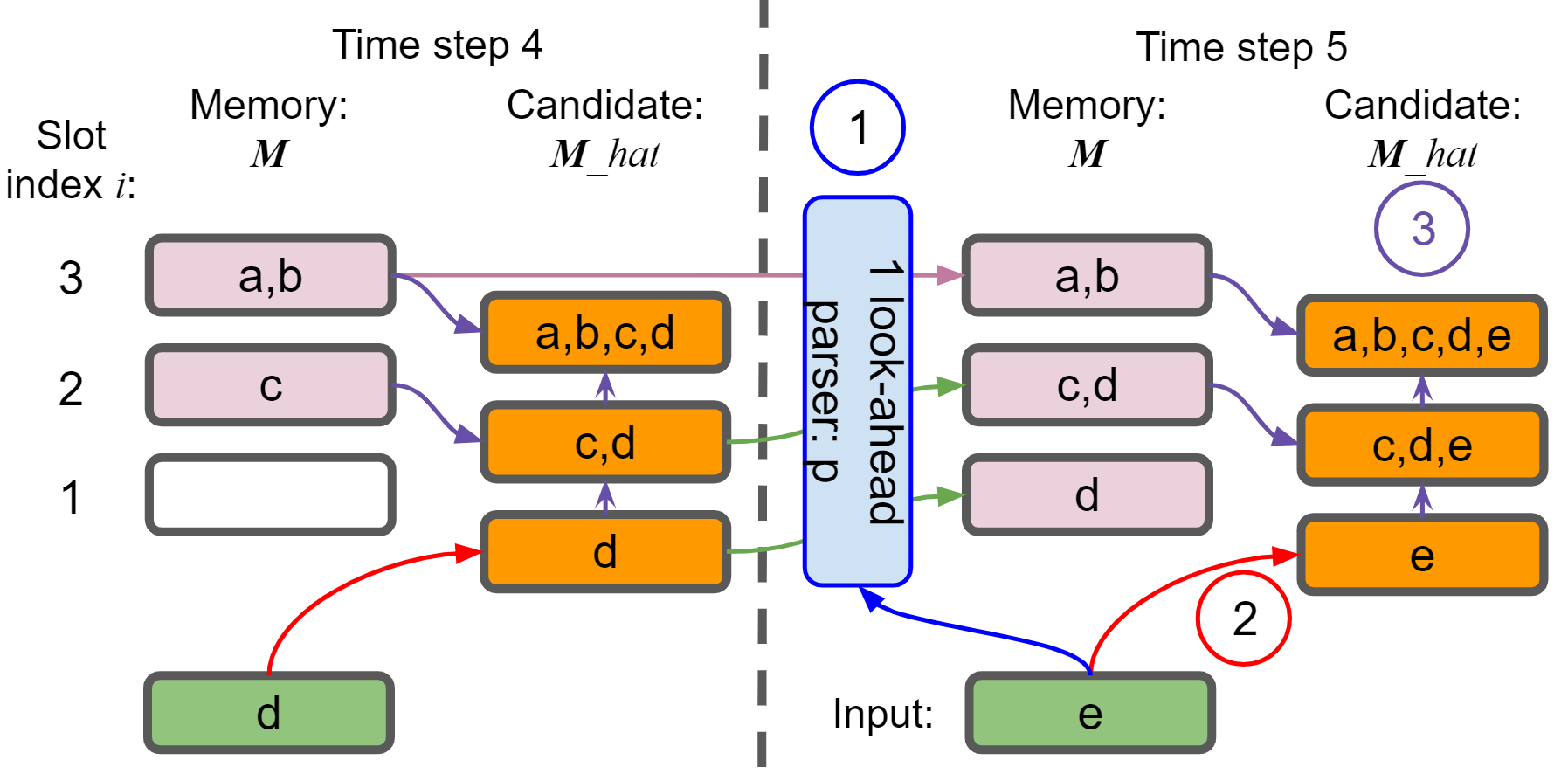}
  \caption{ The transition from time step 4 to 5.
  \circled{1} The one-step look-ahead parser combines $\hat{M}_{t-1}$ and $M_{t-1}$ considering on the current input $x_t$,
  in this example, the split point of $\hat{M}_{t-1}$ and $M_{t-1}$ is $i=2$.
  \circled{2} Current input $x_t$ is written into the lower slot of new candidate memory $\hat{M}_{t}^{i-1}$. 
  \circled{3} The rest of new candidate memories $\hat{M}_{t}^{\geq i}$ are generated with bottom-up recurrent composition.
  }
  \label{fig:om_timestep}
\end{subfigure}
\quad
\begin{subfigure}{.45\textwidth}
 \centering
    \includegraphics[width=1\linewidth]{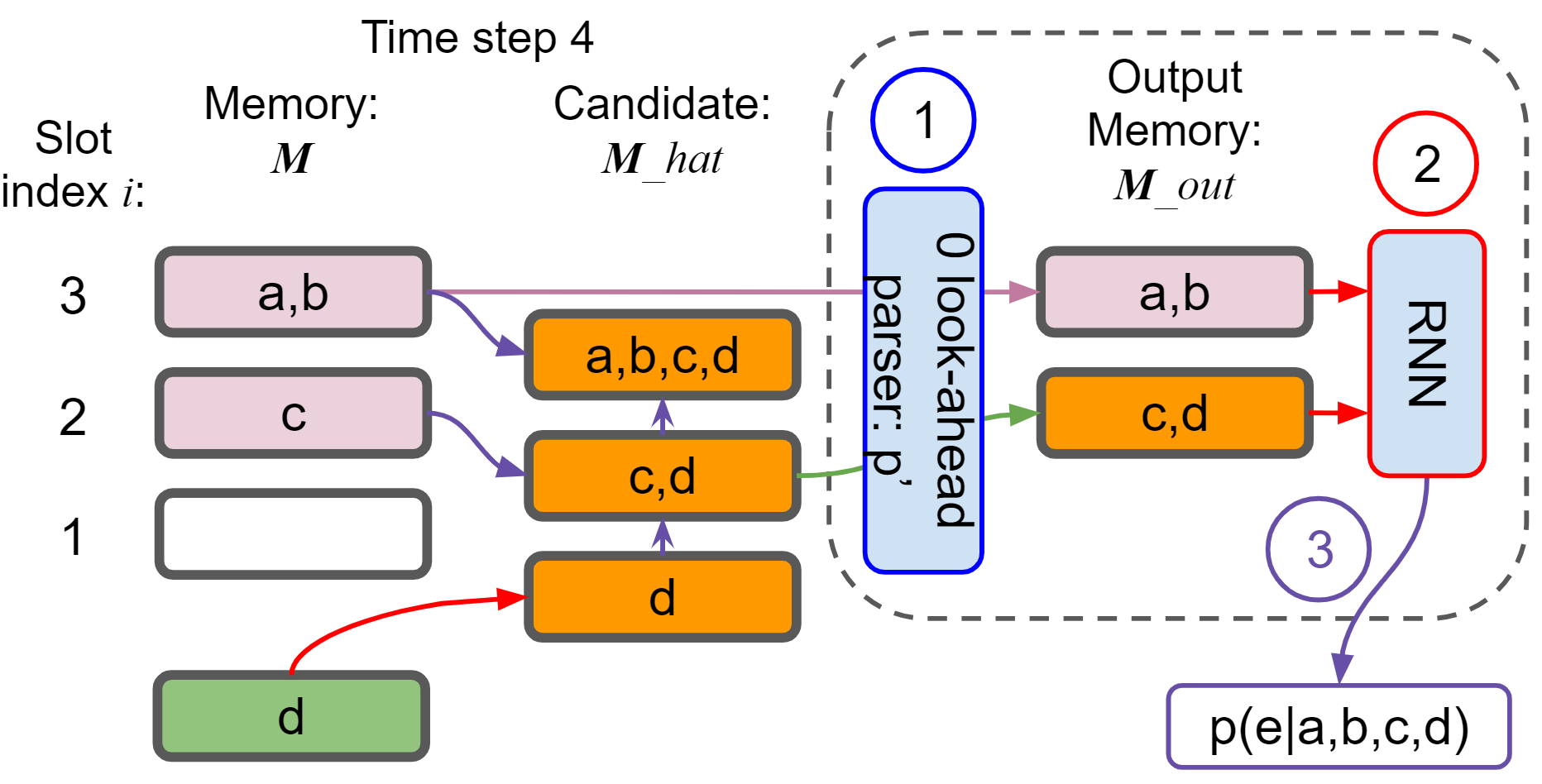}
    \caption{ Predicting the next token at time step 4. 
    \circled{1} The zero-step look-ahead parser combines $M_t$ and $\hat{M}_t$ at time step $t$.
    \circled{2} The recurrent network takes the combined memory $M^{\mathrm{out}}_t$ as input and output a hidden state $h_t=f(w_{\leq t})$.
    \circled{3} $h_t$ is then fed into an linear layer to compute $p(x_{t+1}|x_{\leq t})$.
    }
    \label{fig:decoder}
\end{subfigure}
\caption{ The recurrent transition (left) and prediction network (right) of SOM. 
In the recurrent transition, a one-step look-ahead parser predict the syntax once \emph{e} is observed and can be seen as a posterior over the syntax given the current word.
The prediction network uses a zero-step look-ahead parser to predict the location of the next phrase and acts as a prior on the syntactic structure.}
\label{fig:ordered_memory}
\vspace{-0.3cm}
\end{figure*}

We first provide useful background on Ordered Memory. Ordered Memory (OM, \citealt{shen2019ordered}) is a recurrent neural network that explicitly models recursive structure through memory writing and erasing operations.
OM maps the latent syntax into a $T \times N$ memory grid $\Tilde{M}$, where $T$ is the length of input sequence and $N$ is the maximum number of memory slots. 
Figure~\ref{fig:grid_view} gives an intuition of what the grid contains.
Empty blocks in the figure represent memory slots that can be discarded during inference.
Ideally, the memory network should generate the $t$-th column of the grid $\Tilde{M}_t$ at time step $t$.
But generating $\Tilde{M}_t$ requires the model to have access about the tree structure which is usually latent. For this reason, OM induces the latent structure through inductive biases of its reading and writing operations.

As a recurrent model, OM performs one-step look-ahead incremental parsing through maintaining three states:
\begin{itemize}
    \item \textbf{Memory $M_{t}$}: a matrix of dimension $N \times D$, 
    where each occupied slot is a distributed representation for a node spanning an subsequence in $x_1, .., x_{t-1}$ conditioned on $x_t$, i.e. $M_t$ represents a one-step look-ahead parser stack. It's represented by gray blocks in Figure~\ref{fig:ordered_memory}.

    \item \textbf{Candidate memory $\hat{M}_{t}$}: a matrix of dimension $N \times D$ contains representations for all possible new nodes at time step $t$. 
    At next time step $t+1$, the model will decide whether or not to write these candidates into memory $M_{t+1}$ conditioned on $x_{t+1}$.
    They are represented by orange blocks in Figure~\ref{fig:ordered_memory}. if the model is making correct parsing decisions, then $M_t = \Tilde{M}_{t-1}$.
    
    \item \textbf{Memory mask $\cp_{t}$}: $\cp_{t} \in \{0, 1\}^N$, where each entry indicates whether the respective slot in $\hat{M}_{t}$ is occupied by a candidate,~e.g., if $\cp_{t} = (0, 1, 1)$, then the occupied slots are $\hat{M}_t^{\geq 2}$. At next time step, the model can only choose a candidate from masked slots to write into the memory $M_{t+1}$.
\end{itemize}
At each time step, the model takes $[ M_{t-1}, \hat{M}_{t-1}, \cp_{t-1} ]$ and word embedding $x_t$ as inputs, returning the outputs $[M_{t}, \hat{M}_{t}, \cp_{t}]$. 

To generate the new memory $M_t$, we combine $M_{t-1}$ and $\hat{M}_{t-1}$ to match $\Tilde{M}_{t-1}$.
The model uses $x_t$ as its query to attend on previous candidates $\hat{M}_{t-1}$. 
The attention distribution is $p_t$, which models the split point of gray blocks and orange blocks in Figure~\ref{fig:grid_view}. 
Suppose $p_t$ is a one-hot distribution and $p_t^i=1$.
The candidates $\hat{M}_{t-1}^{\leq i}$ are written into the respective memory slot $M_t^{\leq i}$, while $M_{t-1}^{>i}$ are copied to $M_t^{>i}$:
\begin{align}
    M_t^{\leq i} = \hat{M}_{t-1}^{\leq i}, \quad M_t^{> i} = M_{t-1}^{> i}
\end{align}
We will refer to the process of generating $M_t$ as a one-step look-ahead parser, since the model is using the current input $x_t$ as extra information to build the partial parse for time step $t-1$.
To generate new candidates $\hat{M}_t$, the input embedding $x_t$ is written into $\hat{M}_t^{i-1}$, and $\hat{M}_t^{\geq i}$ are computed recurrently with eq.\ref{eq:recurrent}:
\begin{align}
    \hat{M}_t^{< i - 1} &= \emptyset, \quad \hat{M}_t^{i-1} = x_t \\
    \hat{M}_t^j &= \cell(M_{t}^j, \hat{M}_{t}^{j-1}), & \forall j \geq i \label{eq:recurrent} 
\end{align}
where $\cell()$ is the composition function that takes its childrens' representations as input and output the parent's representation.
The non-empty slots in candidate memory are then $\hat{M}_t^{\geq i-1}$, and they can be masked by:
\begin{align}
    \cp_t^{< i - 1 } = 0, \quad \cp_t^{\geq i - 1} = 1
\end{align}
In other words, $\cp^{i}_t = \sum_{j\leq i+1}p^{j}_t$, and $\cp_{t}^i$ is monotonically increasing.
More details of the OM can be found in \citet{shen2019ordered}.

\section{Syntactic Ordered Memory}\label{sec:model}

We propose two augmentations to OM in order to better perform language modelling and incremental parsing: a prediction network and the dynamic oracle.
a) Previous language models mostly focus on predicting the next token or a missing token. 
In our case, we are explicitly modeling the latent structure.
By predicting the structure for the next token, we exploit this latent structure for word prediction.
This helps the model better organize information for predicting next word, allowing shortcuts to be created for long-term dependencies, as shown in Fig.\ref{fig:high_level}. 
b) If the model only observes states resulting from correct past decisions at training time, it will not be prepared to recover from its own mistakes during prediction, suffering from exposure bias \citep{schmidt2019generalization, fried2018policy}.
In the experiment section, we demonstrate how this phenomenon will significantly hurt the language model performance and, to a lesser extent, also hurt the parsing performance.

\subsection{Prediction Network}

At time step $t$, the prediction network takes $[M_{t}, \hat{M}_t, \cp_{t}]$ as input, and produces a probability distribution over the next token $p(w_{t+1} | w_{\leq t})$.
To do this, we need to have a temporary estimate of the local structure.
We therefore need to approximate $p_{t+1}$ with a zero-step look-ahead prediction $\q_{t}$:
\begin{align}
    \alpha_t^i &= \frac{ \mathbf{w}^{Att}_2 ~ \mathrm{ReLU} \left( \mathbf{W}^{Att}_1  \hat{M}_{t}^i + b_1 \right)+ b_2 }{\sqrt{N}} \\
    \q_t &= \mathtt{masked\_softmax} (\alpha_t, \mathrm{mask} = \cp_t)
\end{align}
where $\mathbf{W}^{Att}_1$ is $N \times N$ weight matrix, $\mathbf{w}^{Att}_2$ is a $N$ dimension weight vector, and $\alpha^i_t$ is a scalar. 
We then sample the slot at index $i$ from the distribution $\q_t$.
$i$ is the zero-step look-ahead parsing decision, which means that the next phrase will be a sibling of node $\hat{M}_t^{i}$.
We therefore need to predict the next token conditioned on $\hat{M}_t^{i}$ and its previous contexts.
So we feed memory slots $[M_t^N, M_t^{N-1}, ..., M_t^{i+1}, \hat{M}_t^{i}]$ into a recurrent neural network:
\begin{equation}
    h_t = \mathtt{RNN} \left( M_t^N, M_t^{N-1}, ..., M_t^{i+1}, \hat{M}_t^{i} \right)
\end{equation}
where $h_t$ is the final hidden state of the RNN. As shown in Figure~\ref{fig:decoder}, the input sequence are representations of non-overlapping subtrees spanning from $x_1$ to $x_t$. $h_t$ can therefore be seen as a distributed representation of the sequence $w_{\leq t}$.
In the RNN, we use the same architecture as the cell function in OM to model the recurrent transition function:
\begin{align}
    \left[
    \begin{matrix}
        f_j \\ 
        i_j \\
        c_j \\
        u_j
    \end{matrix}
    \right] &=  \mathbf{W}_2^{Cell} \mathtt{ReLU} \left( \mathbf{W}_1^{Cell}  \left[ 
    \begin{matrix}
        h_t^{j + 1} \\ 
        M_{j} 
    \end{matrix} 
    \right] 
    + b_1 \right) + b_2 \\
    h_t^j &= \mathtt{LN} (\sigma(f_j) \odot h_t^{j + 1} + \sigma(i_j) \odot M_{j} + \sigma(c_j) \odot u_j )
\end{align}
where $\sigma$ is the sigmoid function, $\mathtt{LN}$ is layer normalization function, $f_j, i_j, c_j$ are controlling gates, $c_j$ is cell state, and $h_t^{N+1}$ is a zero vector. After obtaining $h_t$, we can compute the distribution over the next token and the language modelling loss:
\begin{align}
    p(w_{t+1} | w_{\leq t}) &= \mathtt{softmax} (\mathbf{W}_{emb} h_t + b) \\
    L_{\mathrm{LM}} &= - \sum_t \log( p(w_{t+1} | w_{\leq t}) )
\end{align}

\subsection{Dynamic Oracle for SOM}
\begin{algorithm}
    \SetAlgoLined
    \KwData{$\theta_1, ..., \theta_{T}$, $\Gamma$}
    \KwResult{$\xi_1, ..., \xi_T$}
    initialize $\xi_1=N$\;
    
    \For{$i \leftarrow 2$ \KwTo $T$}{
        $j = \mathtt{first\_sibling}_{\Gamma}(i)$\;
        $\mu_i = \max(\theta_{j+1}, ..., \theta_{i-1})$\;
        $\xi_i = \max(\xi_j - 1, \mu_i)$\;
    }

    \caption{\fontsize{10pt}{12pt}\selectfont
    The structure label generation algorithm, where $\Gamma$ is the ground-truth tree and $\theta_i$ is the structural decisions made by our model.
    This algorithm produces a parse close to the original given the errors already made, and that new gold parse is converted into grid decisions.
    Given $\Gamma$, the function $\mathtt{first\_sibling}_{\Gamma}(i)$ returns the index of the first token in the smallest clause that contains $w_i$, and where $w_i$ is not the first token. 
    Ideally, $w_i$ should be written into the slot $(\xi_j - 1)$. 
    For example, in Figure~\ref{fig:grid_view}, $c$ is written into the slot 2, then $d,e$ should be written into the slot 1.
    However, the model could make a wrong decision between $w_j$ and $w_i$. 
    If the model has merged information from $w_j$ into a higher slot $\mu_i$, $x_i$ should be written into slot $\mu_i$ as well.
    }
    \label{algo:structure_label}
\end{algorithm}

One way to provide a supervision signal for $p_t$ and $\q_t$ is to train the parser with static oracle: feed the gold tree to the model, and have the model predict future decisions.
However, static oracle makes the language model overfit on the gold tree, resulting in bad perplexity scores (Table~\ref{tab:ablation}).
Inspired by the dynamic oracles proposed in \citep{goldberg2012dynamic, coavoux2016neural}, we propose a dynamic oracle for ordered memory, which dynamically changes the reference structure based on mistakes made by our model on previous steps.
To do this, we build the structure label for each time step based on the gold tree and previous decisions made by the model. 
During training, we sample the model's decision from $p_t$:
\begin{equation}
    \theta_t = \mathtt{Multinomial}(p_t)
\end{equation}
and we make greedy decisions during evaluation:
\begin{equation}
    \theta_t = \mathtt{argmax}(p_t)
\end{equation}
The same operations are applied to $\q_t$ as well.

We use the Algorithm.\ref{algo:structure_label} to convert the gold tree $\Gamma$ into labels $\xi_t$ for $p_t$. Since the zero-step look-ahead distribution $\q_t$ should match the one-step look-ahead distribution $p_{t+1}$ at next time step $t+1$, we use $\xi_{t+1}$ as label for $\q_t$.
The structure loss is the negative log-likelihood:

{\small
\begin{equation}
    L_{\mathrm{S}} = -\sum_{t} \left( \log(p_t(\xi_t | w_{\leq t})) + \log(\q_t(\xi_{t+1} | w_{\leq t})) \right)
    \nonumber
\end{equation}
}

\begin{table}[]
    \centering
    \small
    \begin{tabular}{c|c|c|c}
        \hline
        Type & Max & Median & Mean \\
        \hline
        Constituency & 29 & 7 & 7.7 \\
        Dependency & 16 & 4 & 4.2 \\
        \hline
    \end{tabular}
    \caption{ Statistics of tree depth for Penn Treebank. Dependency trees are converted from constituency tree with Stanford Corenlp toolkit.}
    \vspace{-0.3cm}
    \label{tab:tree_depth}
\end{table}

\begin{figure}
    \centering
    \includegraphics[width=1\linewidth]{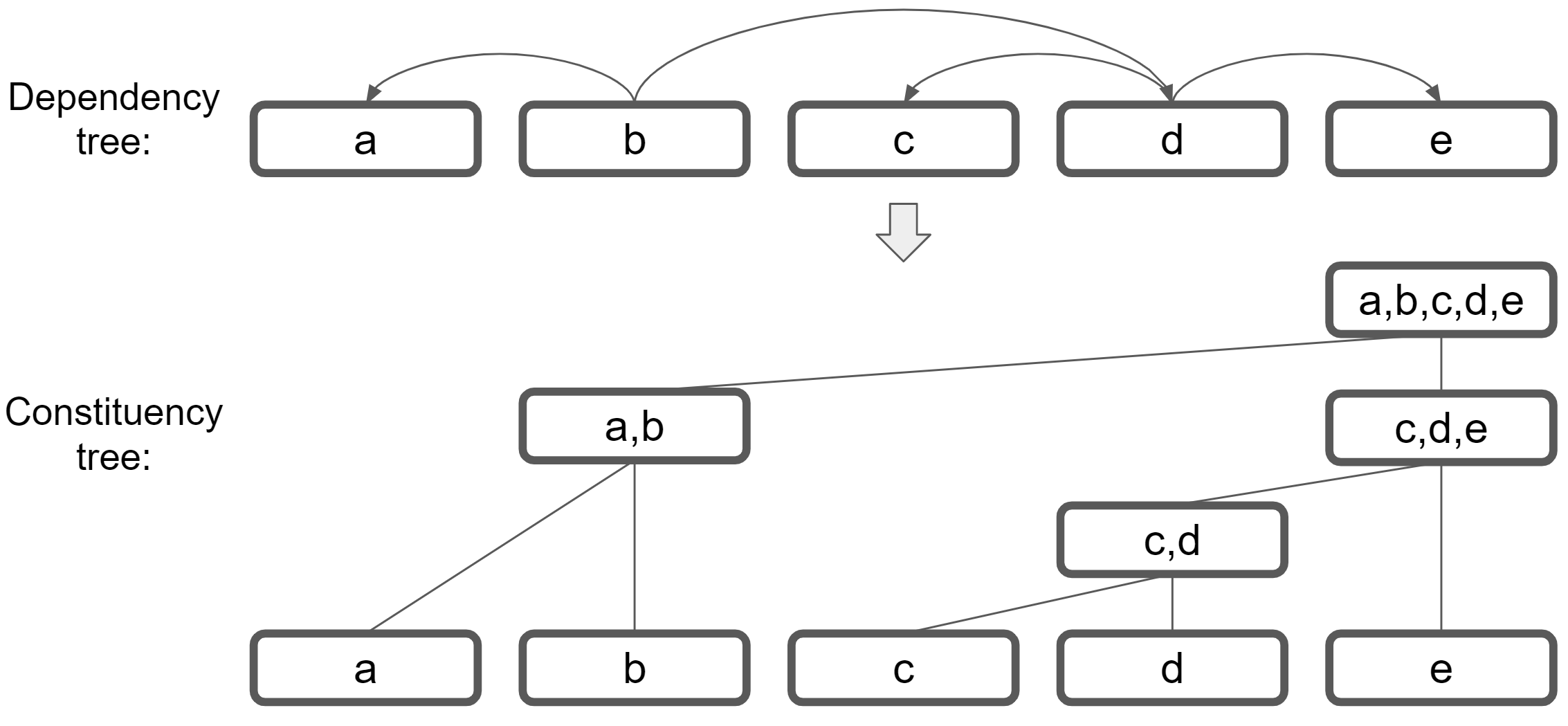}
    \caption{ 
    The universal dependency tree is converted into a constituency tree $\Gamma$ through merging the head and its children into one single constituent.
    Since the grid view only works with binary trees, we binarize n-ary nodes with a left branching bias.
    }
    \vspace{-0.3cm}
    \label{fig:tree_converte}
\end{figure}

For our model, the depth of $\Gamma$ has a linear relation to the computational complexity and GPU memory consumption.
To maximize the model's efficiently, the gold tree $\Gamma$ is constructed from \emph{universal dependency trees}.\footnote{\url{https://universaldependencies.org/}}
There are two reasons we chose universal dependency trees instead of constituency trees: 1) In Table~\ref{tab:tree_depth}, the dependency trees are on average shallower than constituency trees; this means faster computation time and less memory consumption for our model. 2) Universal dependency trees can be applied to many more languages than Penn Treebank-style constituency grammar. Additionally, Penn Treebank-style trees can easily be converted to universal dependency trees.
As shown in Figure~\ref{fig:tree_converte}, we convert the universal dependency tree into $\Gamma$ by merging the head and its children into one single constituent.

\section{Experiments}\label{sec:experiments}
We present the results of SOM on language modeling, syntactic generalization, and incremental parsing.
Details of hyperparameters and experiment settings can be found in Appendix~\ref{app:hyperparameters}.

\subsection{Language Modeling}
{\bf Penn Treebank} has one million words of 1989 Wall Street Journal corpus annotated with constituency trees.
Since SOM primarily focuses on sentence-level structure and language modeling, we use the same preprocessing schema as RNNG\footnote{2-21 for training, 24 for validation, 23 for evaluation.} \citep{dyer2016recurrent}.
Sentences are modeled separately, punctuation is retained, and singleton words are replaced with the Berkeley parser’s mapping rules\footnote{\url{http://github.com/slavpetrov/berkeleyparser}}, resulting in 23,815-word types.
Orthographic case distinction is preserved, and numbers (beyond singletons) are not normalized.

\begin{table*}[t]
    \centering
    \small
    \begin{tabular}{l c c c c c}
    \bottomrule
        \textbf{Model} & \textbf{\# parameters} &\textbf{ppl} & \textbf{$p$ acc} & \textbf{UF1} & \textbf{$p'$ acc}  \\
    \midrule
        SOM                                                                    & 17.7M & \bf 77.68 & \bf 0.927 & \bf 87.96 & \bf 0.870 \\
        SOM $-$ Prediction network                                             & 13.0M & 83.63 & 0.923 & 87.09 & -- \\
        SOM $-$ Prediction network $-$ Language Modeling Loss                                            & 13.0M & -- & 0.925 & 86.26 & -- \\
        SOM $-$ Dynamic Oracle $+$ Static Oracle    & 17.7M & 129.27 & 0.913 & 86.58 & 0.849 \\
        SOM $-$ Dynamic Oracle $+$ Left-branching Oracle                & 17.7M & 82.01 & -- & -- & -- \\
        \midrule
        \multicolumn{2}{l}{\small \emph{Inference with External Trees}} \\
        SOM $-$ Predicted tree $+$ Gold tree & 17.7M & 60.87 & 0.947 & 100.00 & 0.884 \\
    \bottomrule
    \end{tabular}
    \caption{ Ablation tests on the PTB dataset. 
    ``$p$ acc'' and ``$p'$ acc'' are the prediction accuracies of the one-step look-ahead and zero-step look-ahead parsers respectively.
    ``UF1'' is the parsing performance with respect to the converted constituency tree $\Gamma$.
    ``$-$ Prediction network'': this model uses the last candidate memory slot $\hat{M}_t^N$ to predict the next token, instead of using the $h_t$ from the prediction network.
    ``$-$ Predicted tree $+$ Gold tree'': the model's parsing decisions were replaced with ground truth decisions; these results can be considered as the performance upper bound of SOM. 
    }
    \label{tab:ablation}
    \vspace{-0.3cm}
\end{table*}

\begin{table}[t]
    \centering
    \small
    \begin{tabular}{lr}
    \toprule
        \textbf{Model} & \textbf{PTB} \\
    \midrule
        \multicolumn{2}{l}{\small \emph{Without annotations}} \\
        RNNLM & 93.2 \\
        PRPN~\citep{shen2017neural} & 96.7 \\
        URNNG~\citep{kim2019unsupervised} & 90.6 \\
        \midrule
        \multicolumn{2}{l}{\small \emph{With annotations}} \\
        RNNG~\citep{dyer2016recurrent} & 88.7 \\
        RNNG $\rightarrow$ URNNG~\citep{kim2019unsupervised} & 85.9\\
        SOM & \bf 77.7 \\
    \bottomrule
    \end{tabular}
    \caption{Perplexities on Penn Treebank datasets. 
    \emph{With annotations} are models that use the gold tree as supervision signal during training.
    Baseline results are from \citet{kim2019unsupervised}}.
    \label{tab:PTB}
    \vspace{-0.3cm}
\end{table}

\begin{table}[t]
    \centering
    \small
    \begin{tabular}{l r r r}
    \toprule
        \textbf{Model} & \textbf{XS} & \textbf{SM} & \textbf{MD} \\
    \midrule
        n-gram & 240.21 & 157.60 & 106.09 \\
        RNNG & 122.46 & 86.72 & 69.57 \\
        LSTM & 98.19 & 65.52 & 59.05 \\
        ON-LSTM & 71.76 & 54.00 & 56.37 \\
        GPT-2 & 529.90* & 183.10* & 37.04* \\
        SOM & \bf 70.41 & \bf 51.47 & \bf 31.95* \\
        
    \bottomrule
    \end{tabular}
    \caption{Perplexities on BLLIP datasets achieved by different models. Perplexity scores across training dataset sizes are not strictly comparable for models that use word-level vocabulary. * results are using GPT-2's subword vocabulary.}
    \label{tab:bllip}
    \vspace{-0.3cm}
\end{table}

\bigskip
\noindent {\bf BLLIP} is a large Penn Treebank-style parsed corpus of approximately 24 million sentences.
We train and evaluate SOM on three splits of BLLIP: BLLIP-XS (40k sentences, 1M tokens), BLLIP-SM (200K sentences, 5M tokens), and BLLIP-MD (600K sentences, 14M tokens). 
They are obtained by randomly sampling sections from BLLIP 1987-89 Corpus Release 1.
All models are tested on a shared held-out tested set.

Following the settings provided in \citep{hu2020systematic}, datasets are preprocessed into two different versions. 
The first setting is similar to the PTB dataset. 
Singleton words are mapped to UNK classes that preserve fine-grained information, such as orthographic case distinctions and morphological suffixes (e.g. \texttt{UNK-ed}, \texttt{UNK-ly}).
The second setting use subword-level vocabulary extracted from the GPT-2 pretrained model rather than the BLLIP training corpora.

\bigskip
\noindent {\bf Results} of language modeling are given in Table~\ref{tab:PTB} and Table~\ref{tab:bllip}. 
SOM consistently outperforms both the annotated model and non-annotated models.
While GPT-2 seems to fail to learn on smaller datasets, SOM still outperforms GPT-2 on the BLLIP-MD dataset with far fewer parameters (34.8M vs 124.4M), and achieves comparable results with the GPT-2 that is trained on a 3 times larger dataset BLLIP-LG \citep{hu2020systematic}.

The ablation test results are shown in Table~\ref{tab:ablation}.
The biggest performance drop comes from replacing the dynamic oracle with static oracle.
We believe that this is due to the model overfitting on the gold tree, and suffering from exposure bias as a result.
Another big performance drop happens after removing the prediction network.
This suggests that predicting the attaching nodes of the next phrase with the zero-step look-ahead parsers helps to predict the next token.
Replacing the gold tree labels with trivial left-branching tree labels also hurts the perplexity. 
This suggests that learning syntactic structure helps language modeling.

\subsection{Syntactic Generalization}
\begin{figure*}
    \centering
    \includegraphics[width=.85\linewidth]{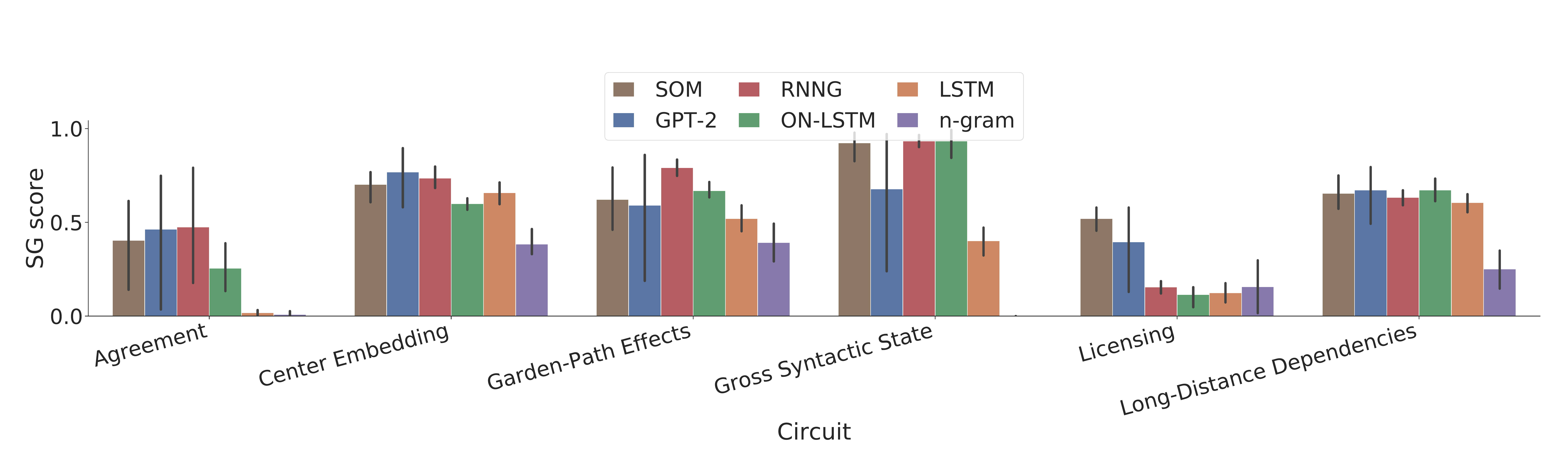}
    \vspace{-0.3cm}
    \caption{ Evaluation results on all models, split across test suite circuits.}
    \label{fig:fine_grained_SG}
    \vspace{-0.3cm}
\end{figure*}
Syntactic Generalization (SG) test suites evaluate the syntactic knowledge of neural language models. 
\citet{hu2020systematic} proposed a set of 34 test suites to evaluation 6 different aspects of syntax: 1) agreement, 2) licensing, 3) garden-path effects, 4) gross syntactic expectation, 5) center embedding, 6) long-distance dependencies.

Following their settings, we evaluate our language models trained on the BLLIP datasets. 
Language models are presented with a group of sentences with minor differences.
To pass each test, the model needs to assign higher conditional probabilities to designated phrases in the sentence that are more grammatical.

    

\begin{figure}
    \centering
    \includegraphics[width=.8\linewidth]{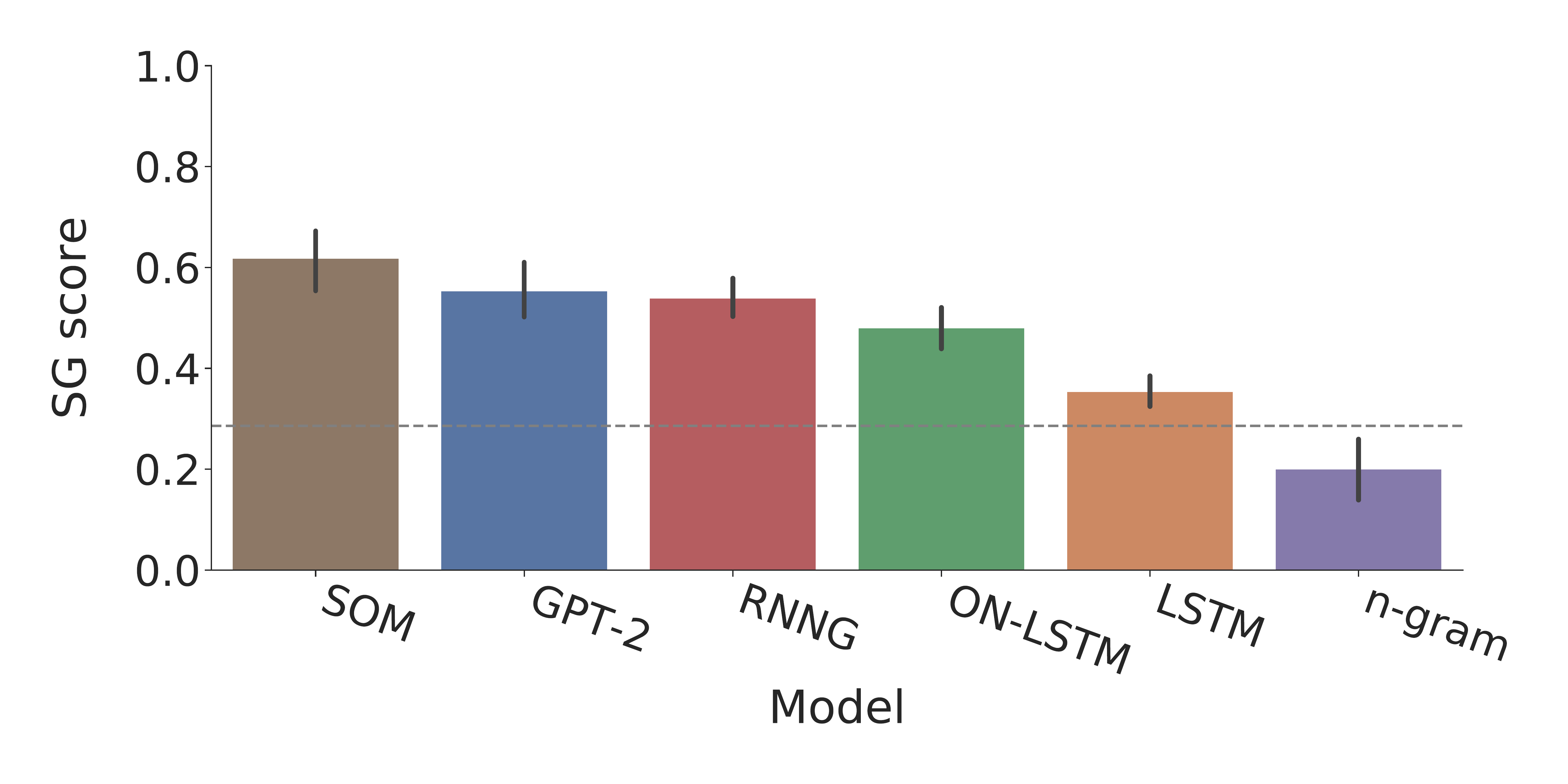}
    \vspace{-0.3cm}
    \caption{ Average SG accuracy by model class.}
    \label{fig:average_SG}
    \vspace{-0.5cm}
\end{figure}

Figure~\ref{fig:average_SG} shows the average accuracy over all model on the complete set of SG test suites.
SOM achieves the best average accuracy, outperforms models with hierarchical structure bias (RNNG, ON-LSTM), and transformer-based model (GPT-2). 
However, according to Figure~\ref{fig:ppl_vs_SG} in Appendix \ref{app:sg}, GPT-2 trained on BLLIP-LG and BLLIP-MD still outperform SOM. 
This could due to that the number of parameters in SOM is largely falling behind GPT-2. 

Figure~\ref{fig:fine_grained_SG} provides fine-grained results on six SG classes.
SOM achieves strong performance on licensing, gross syntactic state, center embedding, and long-distance embeddings.
These classes require the model to keep track of syntactic features across large syntactic chunks (e.g., relative or subordination clauses). 
SOM can effectively keep this long-term information in higher-level memory slots, and revisit the information after the clause in the middle is ended.
More detailed results can be found in Appendix \ref{app:sg}.

\subsection{Incremental Parsing}

\begin{table}[h]
    \centering
    \small
    \begin{tabular}{l c}
    \toprule
        Model & \bf UF1 \\
    \midrule
        PRPN* & 41.2 \\
        ONLSTM* & 47.7 \\
        ONLSTM-SYD \citep{du2020exploiting} & 61.3 \\
        Incremental Shift-reduce Parser & 56.82 \\
        Shift-reduce + LM + Dynamic Oracle & 58.04 \\
        SOM & \bf 67.27 \\
    \midrule
        Oracle Binary Trees & 82.5 \\
    \bottomrule
    \end{tabular}
    \caption{Incremental parsing results on the standard PTB constituency trees. 
    ``*'' means that the model is doing unsupervised grammar induction.
    Since we compare UF1 against the standard, nonbinarized trees (per convention), UF1 scores is upper bounded by the oracle binary trees score.
    }
    \label{tab:constituency_parsing}
\end{table}

To evaluate SOM's performance on incremental parsing, 
we trained and evaluated our models on the standard PTB constituency trees.
Baseline models include:
a) a standard incremental shift-reduce parser with one-step look-ahead; 
b) a incremental shift-reduce parser that equipped with our prediction network and trained on same dynamic oracle and language model loss as our model; 
c) a recently proposed ONLSTM-SYD model \citep{du2020exploiting} that is also trained on both language model and parsing loss; 
d) unsupervised ONLSTM; 
e) unsupervised PRPN.
As shown in Table \ref{tab:constituency_parsing}, SOMs outperform all baseline models, including the shift-reduce parser that has the same extra components as SOMs.
For language modelling performance, original constituency tree based models achieve similar perplexity as dependency tree based counterparts. 
But constituency tree based models require 2$\times$ GPU time and memory to train and evaluate.

For ablation test, we also compare parsing results given by SOM with binary constituency trees $\Gamma$ converted from universal dependency trees.\footnote{UF1 scores are computed by EVALB \url{https://nlp.cs.nyu.edu/evalb/}}
These results are shown in Table~\ref{tab:ablation}.
We observe that using static oracle instead of dynamic oracle results in the worst parsing performance. 
This suggests that our dynamic oracle helps the model to learn a better parser.
After removing the language model loss, the UF1 drops 1.7 points. 
This suggests that the language model loss helps the model to learn better representations for syntax.

\section{Conclusion}
In this work, we propose a new language model with an integrated incremental parser. 
This was done by augmenting the Ordered Memory model with a prediction network, and by using a dynamic oracle for training it to perform incremental parsing.
The resulting model models the joint distribution of syntactic structure and sequence words.
We find that by using the dynamic oracle and explicitly modeling the syntax, we can achieve strong performance on language modelling and syntactic generalization and both these techniques are crucial in the model's performance.

\bibliography{anthology,reference}
\bibliographystyle{acl_natbib}

\clearpage
\appendix

\section{Disentangling Semantic and Syntactic representations}

\begin{figure}[h]
    \centering
    \includegraphics[width=0.7\linewidth]{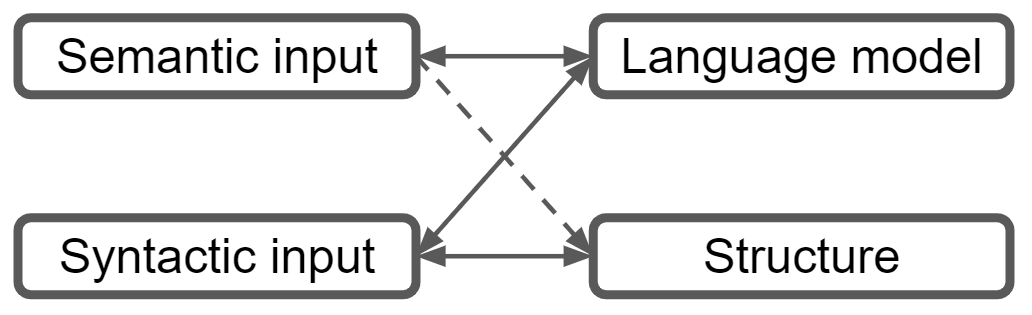}
    \caption{ The schema of disentangling syntax from the language model. 
    Solid lines represent dependency during inference, and gradients flow back during backpropagation.
    The dashed line represents the dependency during inference, but detached so that the gradients do not flow back during backpropagation.
    }
    \label{fig:disentanglement}
    \vspace{-.3cm}
\end{figure}

Given the architecture of our model, we can easily disentangle the language model information flow and parsing information flow.
Figure~\ref{fig:disentanglement} illustrates the disentangled information and gradient flows in our model. 
The language model depends on both prior context and structural inputs, and derivatives are computed with respect to both of these inputs and backpropagated.
However, while the structure also depends on both inputs, we limit backpropagation so that it can only update with respect to the syntactic input.
This is because we want the parsing component to function independently of the language modelling component, but still leverage the semantic information to deal with syntactic ambiguity.

It is possible that existing model architectures could implicitly learn to split these representations, even without the explicit disentanglement that we proposed here.
Yet, Table~\ref{tab:ablation} shows that entangled model can actually achieve stronger in-domain performance, thanks to the liberty to allocate capacity to the two different functionalities based on the training set.

To do so, we propose splitting word embeddings, memory slots, and intermediate hidden states into two segments: semantic segment and syntactic segment.
We then replace linear layers in our cell functions with the following function:
\begin{equation}
    \left[
    \begin{matrix}
        y_{sem} \\ 
        y_{syn} \\
    \end{matrix}
    \right] 
    =
    \left[ 
    \begin{matrix}
        \mathbf{W}_{sem2sem} & \mathbf{W}_{syn2sem} \\ 
        0 & \mathbf{W}_{syn2syn}  
    \end{matrix} 
    \right]
    \left[
    \begin{matrix}
        x_{sem} \\ 
        x_{syn} \\
    \end{matrix}
    \right]
    \nonumber
\end{equation}
where $y_{sem}$ and $x_{sem}$ are the semantic segment which is optimized to minimize language modeling loss, $y_{syn}$ and $x_{syn}$ are the syntactic segment which is optimized to minimize both parsing and language modeling loss.
This architecture results in the solid lines in Figure~\ref{fig:disentanglement}.
Additionally, layer normalization functions are replaced with two separate functions for the two segments respectively.
Meanwhile, $p_t$ still depends on both semantic and syntactic segment, but the structural loss does not backpropagate into the semantic segment:
\begin{align}
    p_t = f(x_{t,sem}, x_{t,syn}, \hat{M}_{t,sem}, \hat{M}_{t,syn}) \\
    \frac{\partial p_t}{\partial x_{t,sem}} = 0, \quad \frac{\partial p_t}{\partial \hat{M}_{t,sem}} = 0
\end{align}
and the same for $\q_t$:
\begin{align}
    \q_t = f(\hat{M}_{t,sem}, \hat{M}_{t,syn}) \\
    \frac{\partial \q_t}{\partial \hat{M}_{t,sem}} = 0
\end{align}
This gradient detachment is represented by the dash line in Figure~\ref{fig:disentanglement}. 
In the experiment section, the disentangled models are denoted as dSOM, and entangled models are denoted as SOM.
For dSOM, the dimension of semantic and syntactic segments for memory slots are denoted as $D_{sem}$ and $D_{syn}$ respectively.
Among the proposed models, the eSOM has the best performance on the in-domain test sets. 
Appendix~\ref{app:out-of-domain} shows that the dSOM slightly outperforms eSOM in perplexity on out-of-domain test sets.

\section{Hyperparameters}
\label{app:hyperparameters}

\begin{table}[h]
    \centering
    \small
    \begin{tabular}{c|c|c|c}
    \hline
        \textbf{Model} & \textbf{XS} & \textbf{SM} & \textbf{MD} \\
        \hline
        RNNG & 22.8M & 48.4M & 81.1M \\
        LSTM & 13.4M & 30.5M & 52.2M \\
        ONLSTM+AWD & 30.8M & 44.2M & 61.2M \\
        GPT-2 & 124.4M & 124.4M & 124.4M \\
        dSOM & 16.4M & 39.5M & 34.8M \\
        eSOM & 17.8M & 41.4M & 37.9M \\
        \hline
    \end{tabular}
    \caption{Parameter counts for different models}
    \label{tab:parameter}
\end{table}

Dropout is applied before all linear layers in our model.
They all share the same dropout rate, except the dropout before language model output layer has a different rate.
We also applied embedding dropout which randomly set some embedding vectors to 0.
Hyperparameters are chosen based on the perplexity on validation set.

\begin{table*}[]
    \centering
    \small
    \begin{tabular}{c|c|c|c|c|c|c}
    \hline
        Dataset & $D_{sem}$ & $D_{syn}$ & \#slots & embedding dropout & dropout & output dropout \\
        \hline
        PTB & 300 & 100 & 15 & 0.1 & 0.3 & 0.5 \\
        BLLIP-XS & 300 & 100 & 15 & 0.1 & 0.3 & 0.5 \\
        BLLIP-SM & 400 & 100 & 15 & 0.1 & 0.2 & 0.2 \\
        BLLIP-MD-BPE & 400 & 100 & 15 & 0 & 0.1 & 0.1 \\
        \hline
    \end{tabular}
    \caption{Hyperparameters. The hidden size of eSOM models are always the sum of $D_{sem}$ and $D_{syn}$}
    \label{tab:my_label}
\end{table*}

\section{More Experiment Results}

\subsection{Syntactic Generalization results}
\label{app:sg}

\begin{figure*}[h]
    \centering
    \begin{subfigure}{.45\textwidth}
    \centering
        \includegraphics[width=1\linewidth]{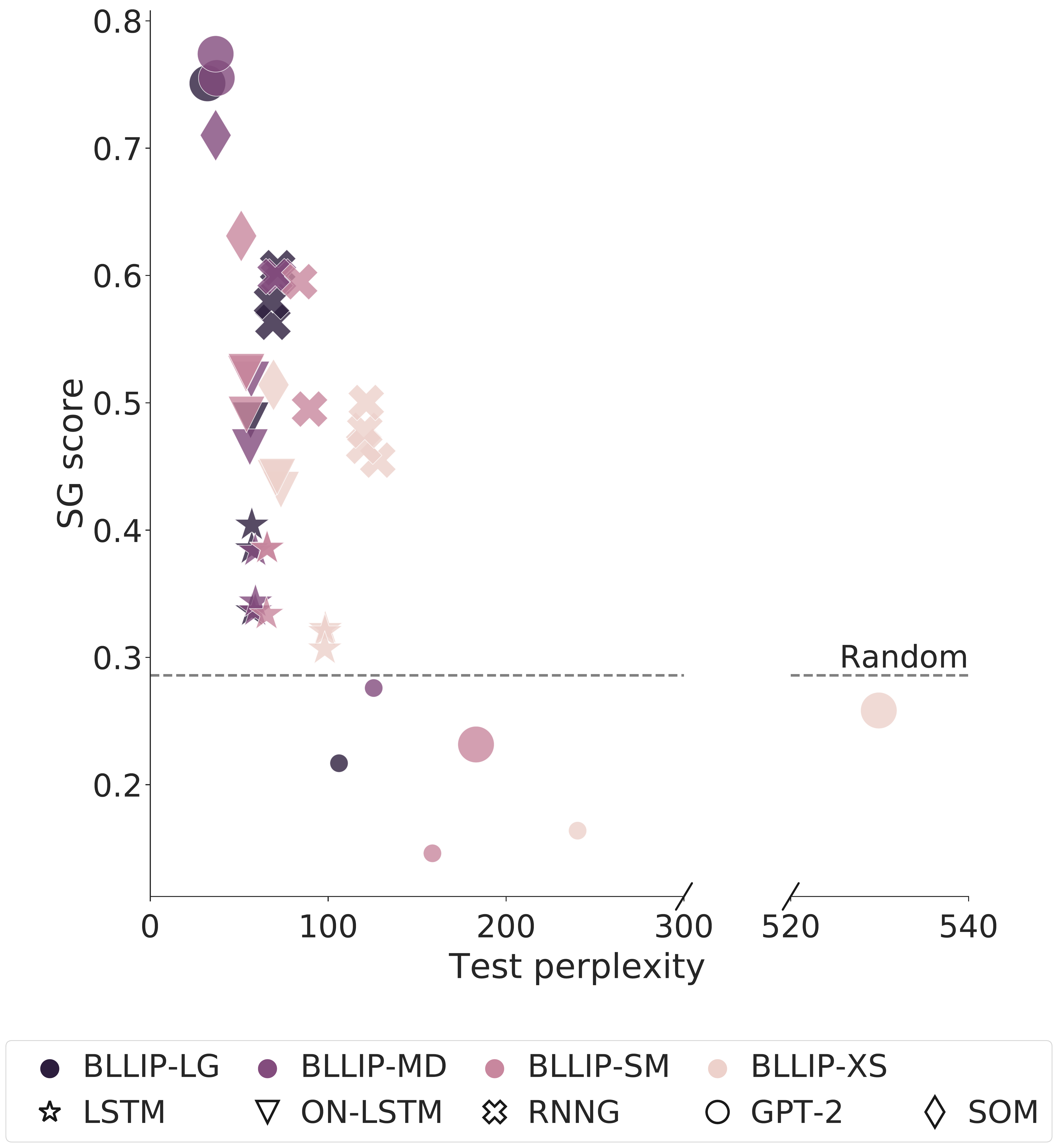}
        \caption{Relationship between SG socre and perplexity on the held-out BLLIP test set.}
        \label{fig:ppl_vs_SG}
    \end{subfigure}
    \quad
    \begin{subfigure}{.45\textwidth}
        \centering
        \includegraphics[width=1\linewidth]{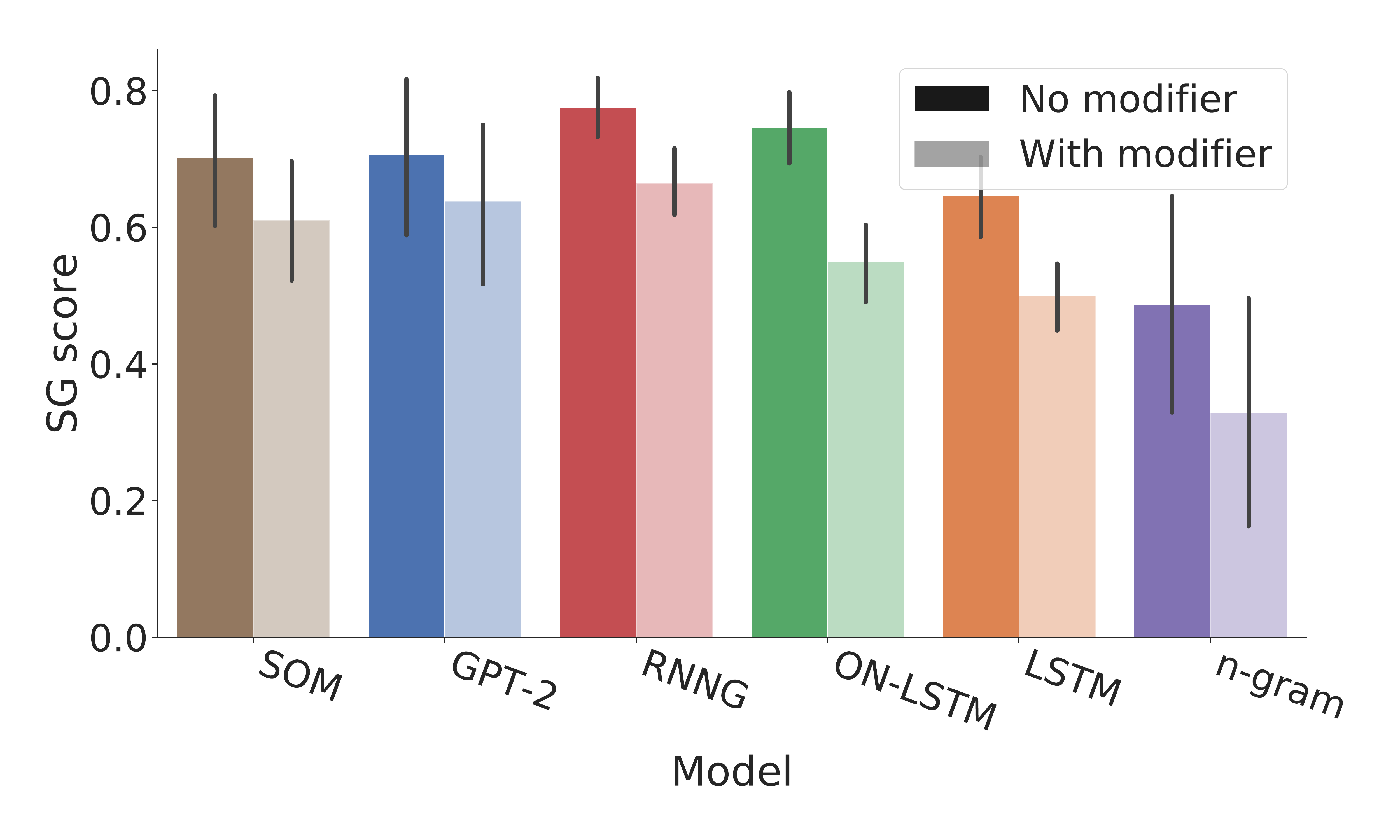}
        \caption{SG score on the pairs of test suites with and without intervening modifiers: Center Embedding, Cleft, MVRR, NPZ-Ambiguous, and NPZ-Object.}
        \label{fig:my_label}
    \end{subfigure}
    \caption{}
\end{figure*}

\begin{figure*}[h]
    \centering
    \includegraphics[width=1\linewidth]{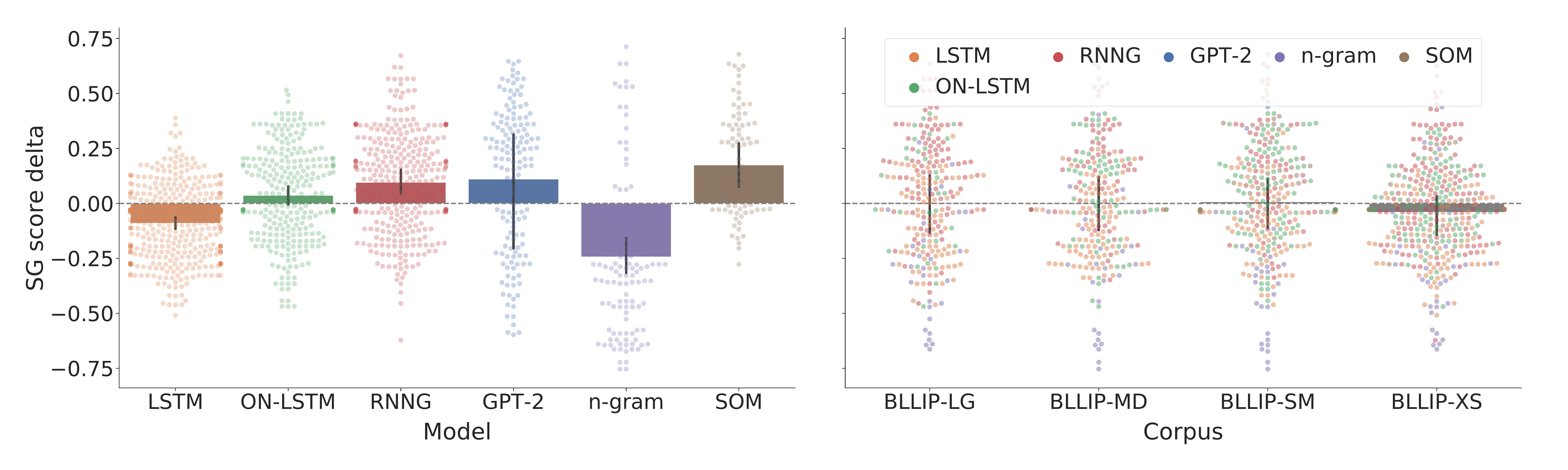}
    \caption{Left: Model class has a trong effect on SG scores. Right: Data scale has little effect on SG scores}
    \label{fig:my_label}
\end{figure*}

\begin{figure*}[h]
    \centering
    \includegraphics[width=1\linewidth]{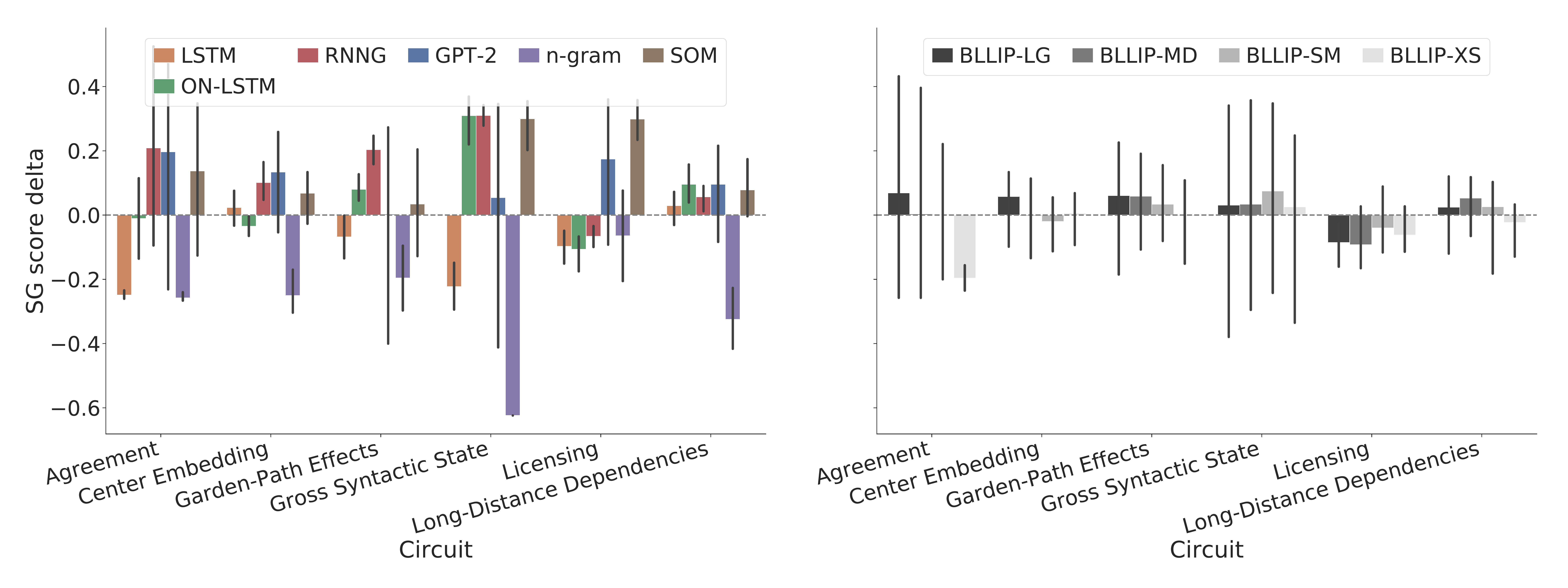}
    \caption{Left: differences in model class induce significant differences in SG scores for several circuits. Right: differences in training data size do not reliably account for most of circuits.}
    \label{fig:my_label}
\end{figure*}

\subsection{Out of Domain Evaluation} \label{app:out-of-domain}

\begin{table*}[h]
    \centering
    \small
    \begin{tabular}{l c c c c c c c c c c c c c}
    \toprule
        \textbf{Dataset} & \multicolumn{2}{c}{GUM} & \multicolumn{2}{c}{EWT} & \multicolumn{2}{c}{ParTUT} & \multicolumn{2}{c}{LinES} & \multicolumn{2}{c}{Pronouns} & \multicolumn{2}{c}{PUD} & SG \\
        \textbf{Metric} & \bf ppl & \bf UF1 & \bf ppl & \bf UF1 & \bf ppl & \bf UF1 & \bf ppl & \bf UF1 & \bf ppl & \bf UF1 & \bf ppl & \bf UF1 & \bf acc\\
    \midrule
        eSOM    & 351.5 & \bf 67.0 & \bf 400.7 & \bf 73.1 & 282.6 & \bf 76.5 & 253.3 & \bf 67.2 & \bf 552.1 & 87.1 & 281.0 & \bf 75.6 & \bf 0.614 \\
        dSOM    & \bf 350.3 & 66.4 & 403.0 & 72.3 & \bf 269.8 & 74.3 & \bf 252.1 & 66.6 & 565.8 & \bf 87.3 & \bf 280.3 & 75.1 & 0.581 \\    
        LB      & 375.5 & -- & 436.5 & -- & 300.9 & -- & 267.5 & -- & 620.4 & -- & 300.7 & -- & 0.513 \\
    \bottomrule
    \end{tabular}
    \caption{ Out of domain test results. 
    Models are trained on PTB. 
    The test sets are obtained from English universal dependencies treebank. 
    ``LB'' stands for left-branching tree labels. 
    Thanks to the structure information, our models generalize much better then the left-branching baseline.}
    \vspace{-0.4cm}
\end{table*}

\bigskip
\noindent {\bf Out-of-domain Test set} contains testsets from other English universal dependencies treebanks.
It contains corpora of different genres, including academic, email, blog, fiction, legal, news, etc. 
We use these datasets to test the generalization ability of models that are trained on PTB.

\end{document}